\documentclass[journal]{IEEEtran}
\usepackage{graphicx}
\usepackage{my_symbols}

\newcommand{\ECC}{\texttt{ECC}}
\newcommand{\ECCr}{\texttt{ECC$_{\textsf{R}}$}}
\newcommand{\FW}{\texttt{FW}}
\newcommand{\FWr}{\texttt{FW$_{\textsf{R}}$}}
\newcommand{\RAk}{\texttt{RAk}}
\newcommand{\RAkr}{\texttt{RAk$_{\textsf{R}}$}}

\newcommand{\MLkNN}{\texttt{MLkNN}}

\newcommand{\IBLR}{\texttt{IBLR}}
\newcommand{\BPMLL}{\texttt{BPMLL}}

\newcommand{\DBNecc}{\texttt{DBN$^2_{\texttt{ECC}}$}}

\newcommand{\DBNbpm}{\texttt{DBN$^{3}_{\textsf{bp}}$}}
\newcommand{\eval}[1]{#1}

\begin{document}

\title{Deep Learning for Multi-label Classification}

\author{Jesse~Read, Fernando~Perez-Cruz}

%

\maketitle

\begin{abstract}

In multi-label classification, the main focus has been to develop ways of learning the underlying dependencies between labels, and to take advantage of this at classification time. Developing better feature-space representations has been predominantly employed to reduce complexity, e.g., by eliminating non-helpful feature attributes from the input space prior to (or during) training. This is an important task, since many multi-label methods typically create many different copies or views of the same input data as they transform it, and considerable memory can be saved by taking advantage of redundancy. In this paper, we show that a proper development of the feature space can make labels less interdependent and easier to model and predict at inference time. For this task we use a deep learning approach with restricted Boltzmann machines. We present a deep network that, in an empirical evaluation, outperforms a number of competitive methods from the literature.


\end{abstract}

\section{Introduction}

Multi-label classification is the supervised learning problem where an instance may be associated with multiple labels. This is opposed to the traditional task of single-label classification (\ie multi-class, or binary) where each instance is only associated with a single class label. The multi-label context is receiving increased attention and is applicable to a wide variety of domains, including text, audio data, still images and video, and bioinformatics, \cite{Thesis,MMD,Overview} and the references therein.

The most well-known approach to multi-label classification is to simply train an independent classifier for each label. This is usually known in the literature as the \emph{binary relevance} (BR) transformation, e.g., \cite{MMD,ECC2}. Essentially, a multi-label problem is transformed into one binary problem for each label and any off-the-shelf binary classifier is applied to each of these problems individually. Practically all the multi-label literature identifies that this method is limited by the fact that dependencies between labels are not explicitly modelled and proposes algorithms to take these dependencies into account.

To date, many successful multi-label algorithms have been obtained by the so-called \textit{problem transformation} methods (where the multi-label problem is transformed into several multi-class or binary problems), for example, \cite{Scene,CLR,ECC,RAKEL,IBL}. These methods make many copies of the feature space in memory (or make many passes over it). Most of the highest performing methods also use ensembles, for example with support vector machines (SVMs) \cite{ECC,RAKEL}, decision trees \cite{HMCens}, probabilistic methods \cite{UPM, LEAD} or boosting \cite{BoosTexter,Subspace2}.

That is to say, most competitive methods from the large part of the literature could benefit tremendously from more concise representations of the feature space, relatively much more so than in the singe-label context; the initial investment in reducing the number of feature variables in a multi-label problem is much more likely to offer considerable speed-ups during learning and classification. However, relatively little work in the multi-label literature has considered this approach. 

Using the raw instance data to construct a model makes the implicit assumption that the labels originate from this data and that they can be recovered directly from it. Usually, however, both the labels and the feature variables originate from particular abstract concepts. For example, we generally think of an image as being labelled \texttt{beach}, not because its pixel-data vector is beach-like, but rather because the image itself meets some criteria of our abstract idea of what a beach is. Ideally then, a feature set would include (for example) variables for a grainy surface such as sand or pebbles, and for being adjacent to a (significant) body of water. Hence, it is highly desirable to recover the hidden dependencies and structure from the original concepts behind the learning task. A good representation of these dependencies make the problem easier to learn. 

A \emph{Restricted Boltzmann Machine} (RBM) \cite{DBNbp} learns a layer of hidden features in an unsupervised fashion. This hidden layer can capture complex dependencies and structure from the input space, and represent it more compactly (whenever the number of hidden units is smaller than the number of original feature attributes). The methods we detail in this paper using RBMs offer some interesting benefits to multi-label classification in a variety of domains:
	
	\begin{itemize}
		\item The predictive performance of existing state-of-the-art methods is generally improved.
		\item Many classification paradigms previously relatively uncompetitive in multi-label learning can often obtain much higher predictive performance and become competitive and thus now offer their respective advantages to this context, such as better posterior-probability estimates, lower memory consumption, faster performance, easier implementation, and incremental learning.
		\item The output feature space can be updated incrementally. This not only makes incremental learning feasible, but also means that cost savings are magnified for batch-learners that need to be retrained at intervals on new data.
		\item The model can be built using unlabeled examples, which are typically obtained much more cheaply than labelled examples; especially in multi-label contexts, since examples are assigned multiple labels.
	\end{itemize}

We also stack several RBMs to create two varieties of \textit{Deep Belief Networks} (DBNs). We look at two approaches using DBNs. In a first approach, we learn the final layer together with the labels and use an existing multi-label classifier. In a second approach, we use back-propagation to fine-tune the weights of our neural network for discriminative prediction, and augment this with a second multi-label predictive layer.

We develop a framework to experiment with RBMs and DBNs in a variety of multi-label classification contexts. Within this framework we carry out an empirical evaluation with many different methods from the literature, on a collection of real-world datasets from diverse domains (to the best of our knowledge, this is also the largest and varied collection of datasets analysed with an RBM framework). The results indicate the benefits of this style of learning for multi-label classification.

\section{\label{sec:prior}Prior Work}

Multi-label datasets and classification methods have rapidly become more numerous in recent years, and classification performance has steadily improved. An overview of the most well known and influential work in this area is provided in \cite{MMD,Thesis}.

The binary relevance approach (BR) does not obtain high predictive performance because it does not model dependencies between labels. A number of methods have improved on this predictive performance with methods that do model label dependence.

A well-known alternative is the \emph{label powerset} (LP) method \cite{Overview} which transforms the multi-label problem into single-label problem with a single class, having the powerset as the set of values (i.e., all possible $2^L$ combinations). In LP, label dependencies are modelled directly and predictive performance is greater than BR, but computational complexity is too high for most practical applications. The complexity issue has been addressed in works such as \cite{RAKEL} and \cite{EPS}. The former presents RAkEL (RAndom $k$-labEL sets), an ensemble method that selects $m$ subsets of $k$ labels and uses LP to learn each of these subproblems.

The \emph{classifier chain} approach (CC) \cite{ECC2} has received recent attention, for example in \cite{PCC} and \cite{UPM}. This method employs one classifier for each label, like BR, but the classifiers are not independent. Rather, each classifier predicts the binary relevance of each label given the input space plus the predictions of the previous classifiers (hence the chain).

Another type of binary-classification approach is the \emph{pairwise} transformation method (PW), where a binary model is trained for each \emph{pair} of labels. The predictions result more naturally in a set of pairwise preferences than a multi-label prediction (thus becoming popular in ranking schemes), but PW methods can be adapted to make multi-label predictions, for example \cite{CLR}. These methods performs well in several domains, although their application can easily be prohibitive on many datasets due to its quadratic complexity.



An alternative to problem transformation is \emph{algorithm adaptation}, where a specific single-label method is adapted directly for multi-label classification. 
\texttt{MLkNN} \cite{MLkNN} is a $k$-nearest neighbours method adapted for multi-label learning by voting from the labels found in the neighbours. \texttt{IBLR} is a related method that also incorporates a second layer of logistic regression. \texttt{BPMLL} \cite{BPMLL} is a back-propagation neural network adapted for multi-label classification by having multiple binary outputs as the label variables. 

Processing the \emph{feature space} of multi-label data has already been studied in the literature. \cite{BrazilFS} presents an overview of the main techniques with respect to problem transformation methods. In \cite{LIFT} a clustering-based supervised approach is used to obtain label-specific features for each label. The advantages of this method are reduced where label-relevances are not trained separately, for example in LP methods (which learns all labels together as a single multi-class meta label). In any case, this a meta technique that can easily be applied independently of other preprocessing and learning techniques, such as the one we describe in this paper. 

In \cite{Subspace2} redundancy is eliminated from the learning space of the BR method by taking random subsets of the training space across an ensemble. This work centers on the fact that a standard BR approach considers the full input space for each label, even though only a subset of the variables may be relevant to any particular label. Compressive sensing techniques have also been used in the literature for reducing the complexity multi-label data by taking advantage of label sparsity \cite{PrincipleLabelSpaceTransformation,CompressedSensing}. 

These methods are mainly motivated by reducing an algorithm's running-time by reducing the \emph{number} of feature variables in the input space, rather than learning or modelling the dependencies between them. 
More examples of feature-space reduction for multi-label classification are reviewed in \cite{MMD}.

The authors of \cite{GuoGu} use a fully-connected network closely related to a Boltzmann machine for multi-label classification, using Gibbs sampling for inference. They use this network to model dependencies in the label space for prediction, rather than to improve the feature space. Since this is a fully connected network, it is tractable only for problems with a relatively small number of labels. 

\begin{figure}[h]
    \centering 
	\caption{\label{fig:prior}A network view of various classifiers; the connections among features and labels.}
	\subfloat[\label{tab:a} \texttt{BR}] {
		\includegraphics[width=0.9\columnwidth]{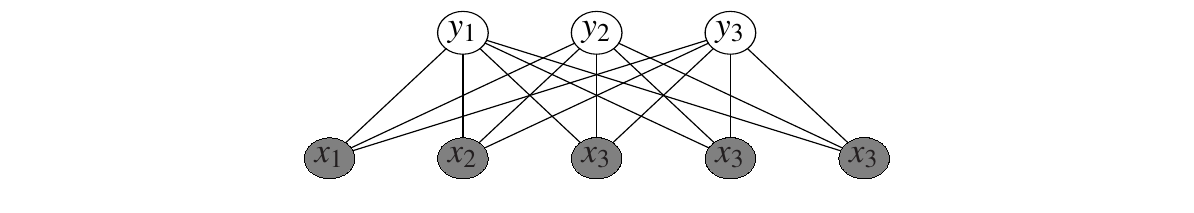}
	}\qquad
	\subfloat[\label{tab:b} \texttt{CC}] {
		\includegraphics[width=0.9\columnwidth]{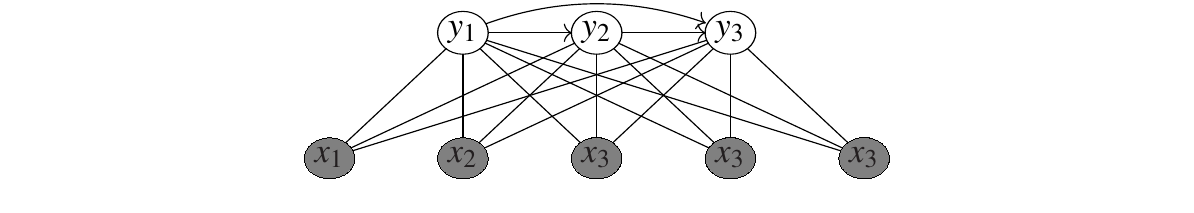}
	}
	\qquad
	\subfloat[\label{tab:c} \texttt{LP}] {
		\includegraphics[width=0.9\columnwidth]{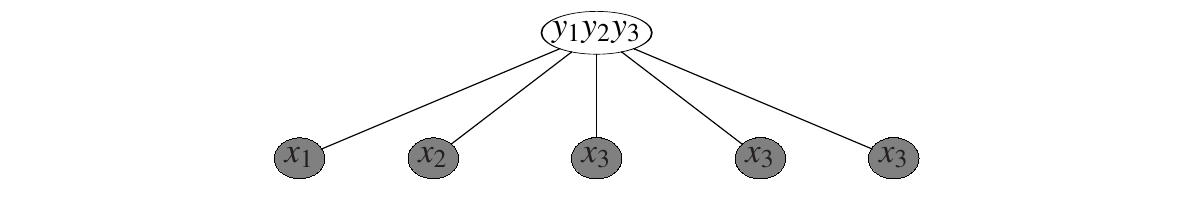}
	}\qquad
	\subfloat[\label{tab:d} \texttt{PW}, \texttt{CDN}] {
		\includegraphics[width=0.9\columnwidth]{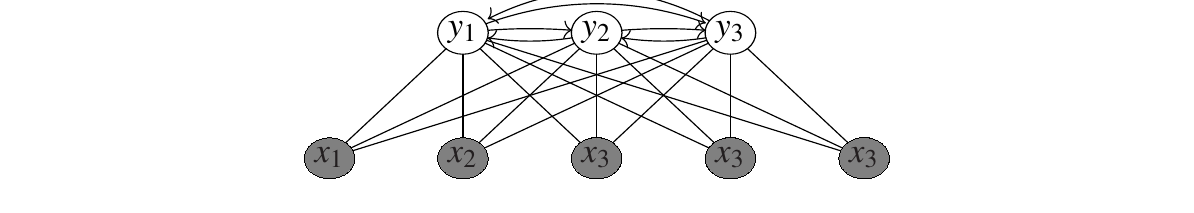}
	}
\end{figure}

\fig{fig:prior} roughly illustrates the way some of the different classifiers model correlations among attributes and labels, assuming a linear base classifier.

\section{\label{sec:back}Deep Learning with Restricted Boltzmann Machines}

A well-known approach to deep learning is to model each layer of higher level features in a \emph{restricted Boltzmann machine} \cite{DBNbp}. We base our approaches on this strategy.

\subsection{Preliminaries}


In all that follows: $X \subset \R^d$ is the input domain of all possible feature values. An instance is represented as a vector of $d$ feature values $\vec{x} = [x_1,\ldots,x_d]$. The set $\L = \{\lambda_1,\ldots,\lambda_L\}$ is the output domain of $L$ possible labels. Each instance $\vec{x}$ is associated with a subset of these labels $Y \subseteq \L$ typically represented by a binary vector $\vec{y} = [y_1,\ldots,y_L]$, where $y_j = 1 \Leftrightarrow \lambda_j \in Y$; i.e., $y_j=1$ if and only if the $j$th label is associated with instance $\vec{x}$, and $0$ otherwise.  

We assume a set of training data of $N$ labelled examples $\{(\x_i,\y_i)\}_{i=1}^N$; $\y_i$ is the label vector (labelset) assignment of the $i$th example; $y^{(i)}_j$ is the relevance of the $j$th label to the $i$th example.

In the BR context, for example, $L$ binary classifiers $h_1,\ldots,h_L$ are trained, where each $h_j$ models the binary problem relating to the $j$th label, such
\begin{align*}
	\ypred &= \h(\xtest) \\
	[y_1,\ldots,y_L] &= h_1(\xtest),\ldots,h_L(\xtest)
\end{align*}
outputs prediction vector $\vec{\hat{y}} \in \{0,1\}^L$ for any test instance $\xtest$. 

\subsection{Restricted Boltzmann Machines}

A Boltzmann machine is a type of fully-connected neural network that can be used to discover the underlying regularities of the (observed) training data \cite{BM}. When many features are involved, this type of network is only tractable in the \emph{restricted} Boltzmann machine setting \cite{DBNbp}, where units are fully connected between layers, but are unconnected \emph{within} layers.

An RBM learns a layer of $u$ hidden feature variables from the original $d$ feature variables of a training set (usually $u<d$). These hidden variables can provide a compact representation of the underlying patterns and structure of the input. In fact, an RBM can capture $2^{u}$ input space regions, whereas standard clustering requires $O(2^{u})$  parameters and examples to capture this much complexity. 

\begin{figure}[h]
    \centering 
	\includegraphics[width=1.0\columnwidth]{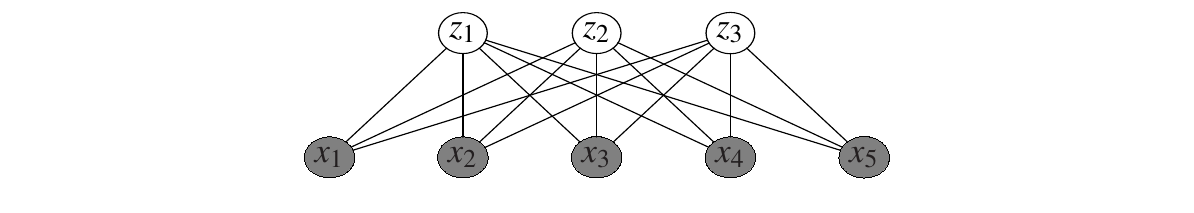}
	\caption{\label{fig:RBM}An RBM with 5 input units and 3 hidden units. Each edge is associated with a weight $W_{jk}$, which together make up weight matrix $\W$. } 
\end{figure}

\fig{fig:RBM} shows an RBM can as a graphical model with two sets of nodes: visible ($X$-variables, shaded) and hidden ($Z$-variables). Each $X_j$ is connected to all $Z_k|k=1,\ldots,u$ by weight $W_{jk}$ (the same for both directions).

RBMs are energy-based models, where the joint probability of visible and hidden units is proportional to the energy between them:
$$
	P(\vec{x},\vec{z}) \propto e^{-E(\vec{x},\vec{z})}.
$$
Hence, by manipulating the energy $E$ we can in turn generate the probability $P(\vec{x},\vec{z})$. Specifically, we minimize the energy
$$
	E(\vec{x},\vec{z}) = -\x\mat{W}\z
$$
by learning the weight matrix $\mat{W}$ to find low energy states. Contrastive divergence \cite{CD} is typically used for this task. 
	
\subsection{Deep Belief Networks}

RBMs can be stacked to form so-called DBNs \cite{DBNbp}. The RBMs are trained greedily: the first RBM takes the input space $\mat{X}$ and produces output $\mat{Z}^{(1)}$, then the second RBM treats $\mat{Z}^{(1)}$ as if it were the input space, and produces $\mat{Z}^{(2)}$, and so on and so forth.

When used for single-label classification, the final output layer is typically a \emph{softmax} function, (which is appropriate where only one of the output units should be on, to indicate \emph{one of} $K$ classes). In the following section we outline our approach, creating DBNs suitable for multi-label classification. 

\section{\label{sec:prop}Deep Belief Networks (DBNs) for Multi-label Classification}



Ideally, an RBM would produce hidden variables that correspond directly to the label variables, and thus we could recover the label vector directly given any input vector; i.e., $\y \equiv \z^{(\ell)}$ or deterministically mappable $\z^{(\ell)} \mapsto \y$. Unfortunately, this is seldom the case, because the abstract hidden variables do not need to correspond directly to the labels. However, we should expect the hidden layer of data to be more closely related to the labels than the original data, and thus it makes sense to use it as a feature space to classify instances. 

Hence, by using the hidden space created by the RBM, we would expect any multi-label classifier to obtain better performance (than when using the original feature space). We do this simply by using the hidden representation of each instance as the input feature space, and associating it with the labels to create training set $\{(\z_i,\y_i)\}_{i=1}^N$. We can then train any multi-label classifier $\h$ on this dataset. To evaluate a test instance $\xtest$, we feed it through the RBM and obtain $\vec{\tilde{z}}$ from the upper layer, and then acquire a prediction
$ \vec{\hat{y}} = \h(\vec{\tilde{z}}) $, and thus so for each test instance. \\





\begin{figure}[h]
    \centering 
	\subfloat[b.1][\label{fig:DBNa}A DBN with two layers of hidden units, i.e., two RBMs.] {
		\includegraphics[width=1.0\columnwidth]{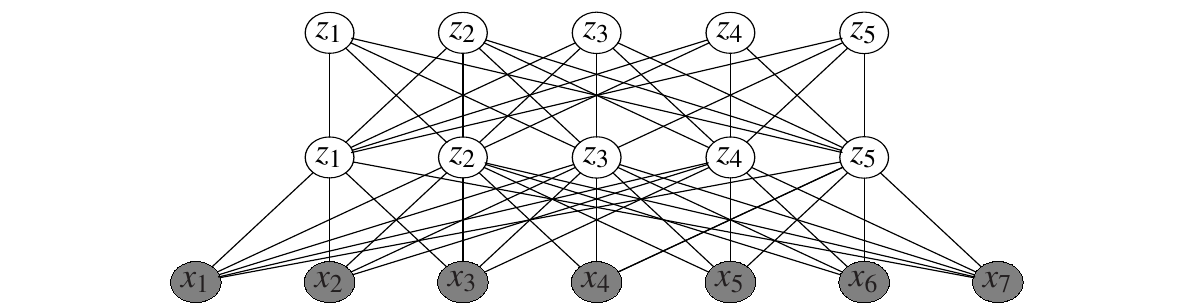}
	}
	\qquad
	\subfloat[b.2][\label{fig:DBNb}A DBN where a 3rd hidden layer represents the labels.] {
		\includegraphics[width=1.0\columnwidth]{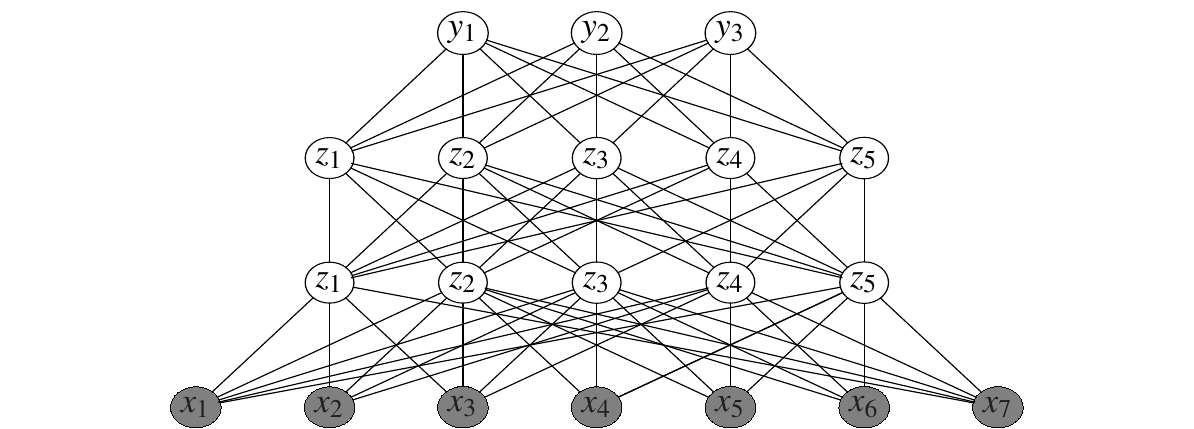}
	}
	\caption{\label{fig:DBN}DBNs for multi-label classification. In \ref{fig:DBNa}, the output space (second hidden layer) $\mat{Z}^{(2)}$ can be trained with the label space $\mat{Y}$ by any multi-label classifier. In \ref{fig:DBNb}, the labels are predicted directly in a third hidden layer.} 
\end{figure}

From here we take two approaches. Since the sub-optimality produced by greedy learning is not necessarily harmful to many discriminative supervised methods \cite{DBNfast}, we can treat the final hidden layer variables $\mat{Z}^{\ell}$ as the feature input variables, and train any off-the-shelf multi-label model $\h$ that can predict
$$
	\ypred = \h(\ztest^{\ell})
$$
where $\ztest^{\ell}$ is produced by the RBM for some test instance $\xtest$; see \fig{fig:DBNa}. 

In a second approach, we add a final layer of weights $\W^{(\ell)}$ on top; see \fig{fig:DBNb}. Now, the structure is similar to the neural network of \texttt{BPMLL} \cite{BPMLL}, except that create the layers and initialize the weights using RBMs. Later we will show that our methods performs much better. We can employ back propagation to fine-tune the network in a supervised fashion (with respect to label assignments) as in, for example, \cite{DBNbp} (for single-label classification). For a number of epochs, each training instance $\x_i$ is propagated forward (upward) through the network and output as the prediction $\ypred_i$. The errors $\boldsymbol\epsilon_i = \y_i - \ypred_i$ are then propagated backward through the network, updating the weights (previously initialized by the RBMs). Due to the initialisation with RBMs, far fewer epochs are required than would usually be typical for back propagation (and we actually observed that more than around $100$ epochs tends to result in overfitting). 


On both these approaches it is possible to add more depth in the form including an additional classification layer. In the multi-label context, this has previously been done to the basic \texttt{BR} method in \cite{2BR}, where a second \texttt{BR} is trained on the outputs of the first (a stacking approach). A related technique in the neural network context, often called a ``skip layer'' has been used in, e.g., \cite{SLNL2, SLNL1}. In our case we allow for generic classifiers. This helps add some further discriminative power for taking into account the dependencies in the label space.  \\

Note that we have also experimented with a DBN that models the instance space and label space together generatively $P(\x,\y,\z)$. In the multi-label setting this complicates the inference, since there are $2^L$ possible $\y$. We tried using Gibbs sampling, but could not obtain competitive results from this model in the multi-label setting compared to our other approaches (even after reducing $\x$ in an RBM first). However, this seems like an interesting direction, and we intend to follow this idea further in future work.

\section{Experiments}

We carry out an empirical evaluation to gauge the effectiveness and efficiency of RBMs and DBNs in a number of different multi-label classification scenarios, using different learning algorithms and a wide collection of databases. We have implemented these methods in the MEKA framework\footnote{\url{http://meka.sourceforge.net}}; an open-source Java-based framework with a number of important benchmark multi-label methods. In this framework RBMs can easily be used in a wide variety of multi-label schemes. The source code of our implementations will be made available as part of the MEKA framework. 

We selected commonly-used datasets from a variety of domains, listed in \tab{table:datasets} along with some basic statistics about them. The datasets vary considerably with respect to the type of data, and their dimensions (the number of labels, features, and examples). In \data{Music}, instances of music are associated with emotions; in \data{Scene}, images belong to categories; in \data{Yeast} proteins may be associated with multiple biological functions, and in \data{Genbase} gene sequences. \data{Medical}, \data{Enron} and \data{Reuters} are text datasets where text documents are associated with categories. These datasets are described in greater detail in \cite{Thesis}. \\

\begin{table}[h]
    \centering 
    \caption{\label{table:datasets}A collection of multi-label datasets and associated statistics, where \eval{LC} is \textit{label cardinality}: the average number of labels relevant to each example.}
	\begin{tabular}{rrrrcccr}
	\toprule
	    		& $N$  &$L$	&$d$		&\eval{LC} &Type 		 \\           
	\midrule           
	Music       &593    &6      &72      &1.87          &audio   \\    
	Scene   	&2407	&6      &294		&1.07          &image	 \\     
	Yeast   	&2417 	&14  	&103	   	&4.24          &biology \\     
	Genbase     &661    &27     &1185    &1.25          &biology \\    
	Medical		&978 	&45     &1449    &1.25          &medical/text	 \\     
	Enron   	&1702	&53     &1001    &3.38          &e-mail/text	 \\     
	Reuters 	&6000	&103    &500		&1.46          &news/text	 \\     
	\bottomrule
	\end{tabular}
\end{table}

\subsection{RBM performance}

We first compare the performance of introducing an RBM, blindly trained, for reducing the input dimension and then try out three of the common paradigms in multi-label classification (namely BR, LP and PW) to test the improvements proposed for this feature extraction algorithm. The RBM would improve the performance of the multi-label classification paradigms, if the extracted features are relevant for better describing the task at hand and will be neutral or negative if those features that have been extracted blindly do not correspond with relevant features for assigning labels. 

The RBM has several parameters that need to be fine-tuned (i.e. number of hidden units, learning rate and momentum) and we use three-fold cross validation to set them. We considered the number of hidden units $u \in \{30,60,120,240\}$, the learning rate $\eta \in \{0.1, 0.01, 0.001\}$, and momentum $\alpha \in \{0.2, 0.4, 0.8\}$. We used weight costs of $2\cdot 10^{-5}$ and $E=1000$ epochs throughout.


\subsubsection{Ensemble of Classifier Chains} 

CC is a competitive BR method that uses the chain rule to improve the prediction for each potential label. As it is unclear what should be the best ordering, we use an ensemble of 50 CC, in which the labels are randomly ordered in each realization (as in \cite{ECC2}). In Table \ref{tab:acc_ecc}, we report the \textit{accuracy}, as defined in \cite{Overview, 2BR, ECC2, EPS}, to report the performance of our multi-label classifiers\footnote{There are a variety of multi-label evaluation measures used in multi-label experiments in the literature; \cite{MMD} provides an overview of some of the most popular. The \textit{accuracy} provides a good balance to gauge the overall predictive performance of multi-label methods \cite{Thesis, ECC2}.}:
\begin{displaymath}
    \eval{accuracy} = \frac{1}{N}\sum_{i=1}^{N} \frac{| \vec{y}_i \wedge \vec{\hat{y}}_i |}{| \vec{y}_i \vee \vec{\hat{y}}_i|},
\end{displaymath}
where $\wedge$ and $\vee$ are the bitwise AND and OR functions, respectively, for $\{0,1\}^L \times \{0,1\}^L \rightarrow \{0,1\}^L$.\\

\begin{table}[h]
	\centering
    \caption{\label{table:0.accuracy} We compare ECC with and without feature extraction using RBMs.}
	\subfloat[\label{tab:acc_ecc} We report the \eval{accuracy} for SVM and logistic regression based multi-label classifiers.] {
		\begin{tabular}{rrrrr}
		\toprule
		 & \multicolumn{2}{c}{SVM} & \multicolumn{2}{c}{Log-Reg} \\
		 \cmidrule{2-3} \cmidrule{4-5}
								&\ECCr      &\ECC       	&\ECCr      &\ECC     \\
		\midrule
			  \data{Music}      & 0.581     & 0.576     &\textbf{0.558}       &0.504    \\
			  \data{Scene}      & \textbf{0.731}     & 0.710      &\textbf{0.709}       &0.554     \\
			  \data{Yeast}      & 0.532     & 0.535        &\textbf{0.513}       &0.504 \\
			  \data{Genbase}    & 0.979     & 0.981      &0.971       &\textbf{0.977}    \\
			  \data{Medical}    & 0.695     & \textbf{0.770}     &0.449       &\textbf{0.706}      \\
			  \data{Enron}      & \textbf{0.469}     & 0.454   &\textbf{0.451}       &0.355        \\
			  \data{Reuters}    & 0.459     & 0.461       &\textbf{0.408}       &0.376    \\
		\bottomrule
		\end{tabular}
	}

	\qquad
	\subfloat[The parameters chosen for \ECCr\ on the first of the two folds (using an internal train/test set of the training set). Parameters for the second fold of each dataset were invariably similar or identical.\label{tab:c}] {
		\begin{tabular}{rrrrrrr}
		\toprule
					& \multicolumn{3}{c}{SVMs} & \multicolumn{3}{c}{Log.\ Reg.} \\
		 \cmidrule{2-4} \cmidrule{5-7}
	                      &  $\eta$&$\alpha$ & $u$  &  $\eta$&$\alpha$ & $u$ \\
		\midrule
	\data{Music}          &  0.1  &0.2  & 120    & 0.1  & 0.8 &  30   \\         
	\data{Scene}         &  0.1  &0.8  & 240    & 0.1  & 0.8 &  60   \\         
	\data{Yeast}         &  0.01 &0.2  & 120    & 0.01 & 0.2 &  30   \\         
	\data{Genbase}       &  0.1  &0.8  & 120    & 0.1  & 0.4 &  60   \\         
	\data{Medical}       &  0.1  &0.6  & 120    & 0.1  & 0.6 &  120  \\         
	\data{Enron}         &  0.1  &0.6  & 120    & 0.1  & 0.6 &  120  \\         
	\data{Reuters}       &  0.1  &0.6  & 120    & 0.1  & 0.6 &  120  \\         
		\bottomrule
		\end{tabular}
	}
\end{table}

In Table \ref{tab:acc_ecc},  \ECCr\ and \ECC, respectively, denote the \eval{accuracy} of the ECC with the RBM-generated features and with the original input space. We have used two different classifiers: nonlinear SVM and logistic regression (linear classifier), both of them have been trained with the default parameters in WEKA. It can be seen that the for the logistic regression classifier the achieved \eval{accuracy} with the generated features by the RBM are significantly better for the \data{Music}, \data{Scene}, \data{Enron}, \data{Reuters} datasets, it only underperforms for the \data{Medical} dataset, and they are comparable for \data{Yeast} and \data{Genbase} datasets. The RBM not only reduces the dimensionality of the input space for the classifier, but it also makes the features suitable for linear classifiers, which allows interpreting the RBM features and understand how each one of them participate in the prediction for each label.  

For the SVM-based ECC classifiers there is not a significant difference when we use the RBM processed features compared to using the raw data directly, as the RBF kernel in the SVM can compensate for the preprocessing done by the RBM. In this case, almost all the results are comparable, except for the \data{Scene} and \data{Medical}, in which, respectively, the \ECCr\ and \ECC\ outperform. We should remark that the linear logistic regression is as good as the nonlinear SVM in most cases, so it seams that using the RBM features reduces the input dimension and makes the classification problem easier, as a linear classifier performs as well as a state-of-the-art nonlinear classifier.

\begin{figure}[h]
    \centering 
	\includegraphics[width=0.90\columnwidth]{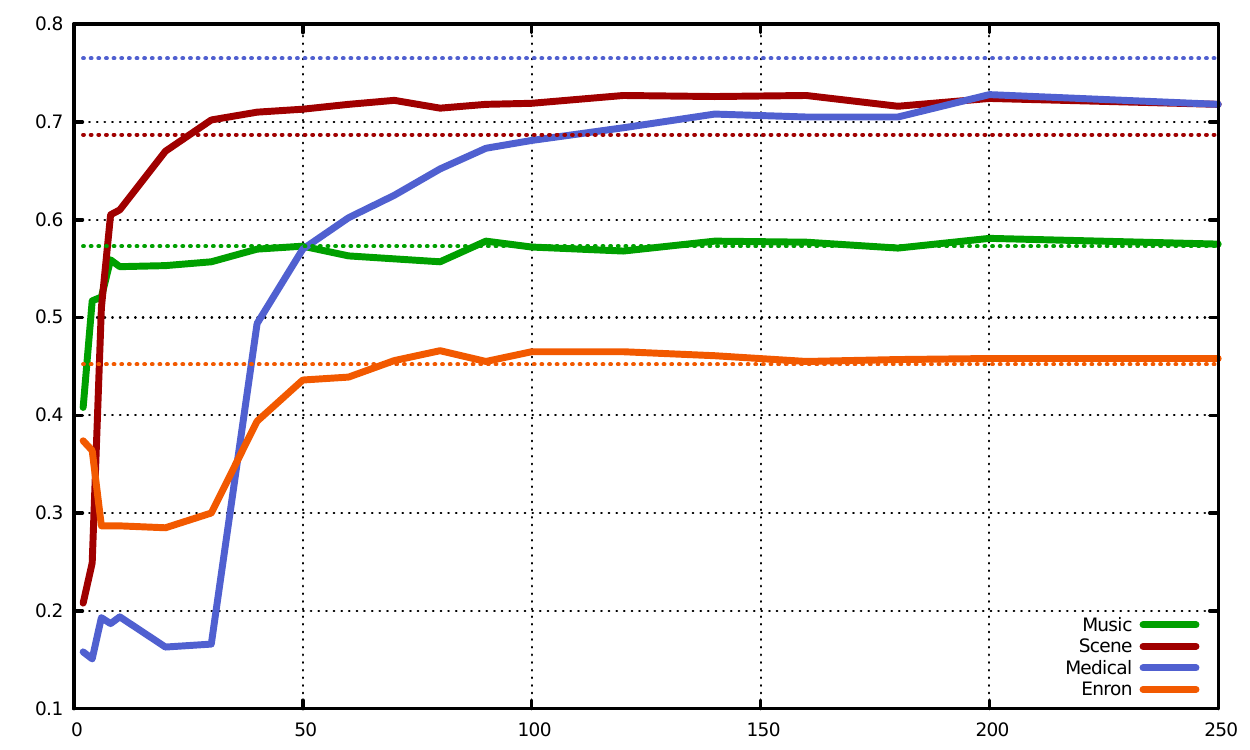}
	\caption{\label{fig:PARAMs} The number of hidden units (horizontal axis) and corresponding accuracy as compared to accuracy with the same methods on the original feature space (horizontal lines). For $\eta=0.1$, $\alpha=0.1$.}
\end{figure}

In Figure \ref{fig:PARAMs} we show the \eval{accuracy} for the seven data bases for the \ECC\ and \ECCr\ multi-label classifier with an SVM classifier, as a function of the number of hidden units of the RBM. In this plot, it can be seen that once we have enough features, using the RBM is comparable to not using it and it is clear that for the \data{Medical} the number of features is too little and we would have needed to increase the number of extracted features\footnote{We did not do so, to keep the experimental setting uniform for all proposed methods, as we think it is important that hyper-parameter setting should be general and not finely tuned for each application.} to achieved the same performance as the SVM does.\\

\begin{figure}[h]
    \centering 
	\includegraphics[width=0.90\columnwidth]{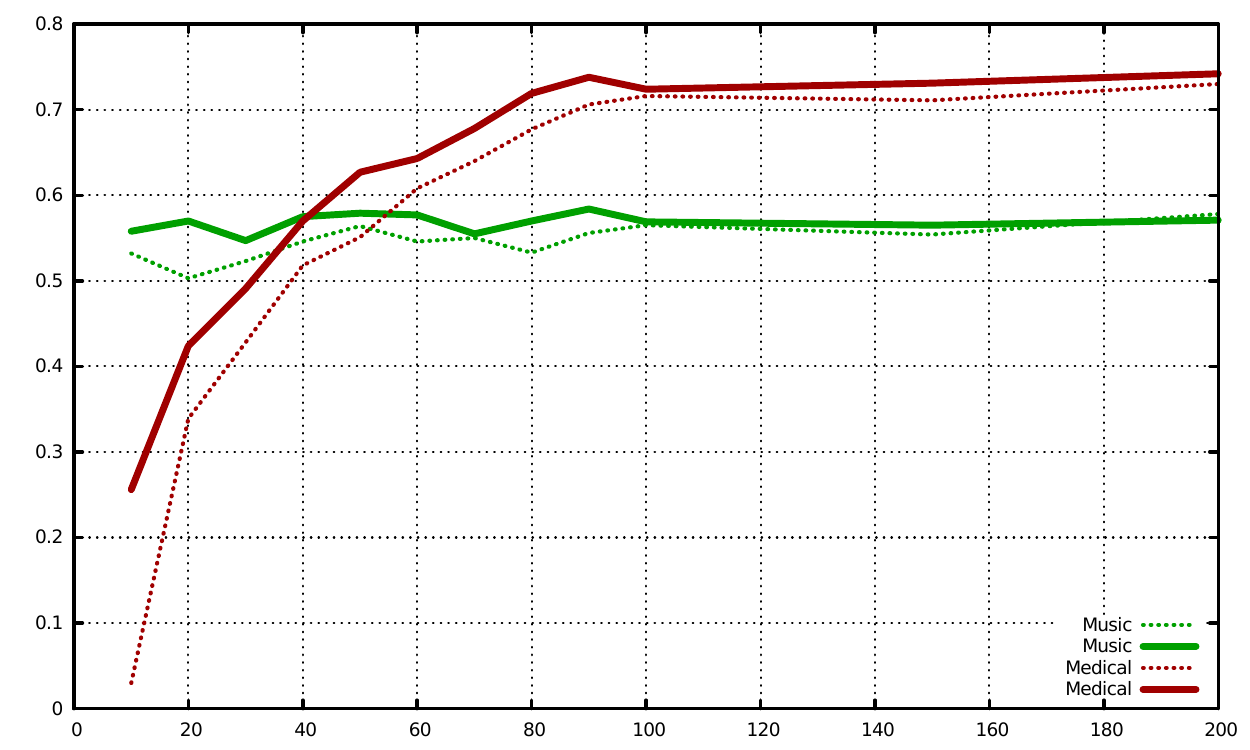}
	\caption{\label{fig:vEBR} The difference in accuracy (shown here on \data{Music} and \data{Medical} datasets) between baseline \texttt{BR} (dashed lines) and more-advanced \texttt{CC} (solid lines) -- both built on RBM-produced outputs -- decreases with more hidden units (horizontal axis). For $\eta=0.1$, $\alpha=0.1$.}
\end{figure}

\begin{table}[h]
	\centering
	\caption{\label{table:new} The accuracy of \texttt{ECC}, with an SVM base classifier, for fixed number of hidden units $u=120$, and for varying learning rate ($\lambda$) and momentum ($\alpha$).}

	\begin{tabular}{lll}
	\toprule
		$\lambda$ & $\alpha$ & \eval{accuracy} \\
	\midrule
		0.001&0.2&0.707\\
		0.001&0.4&0.705\\
		0.001&0.8&0.705\\
		0.01 &0.2&0.710\\
		0.01 &0.4&0.714\\
		0.01 &0.8&0.720\\
		0.1  &0.2&0.726\\
		0.1  &0.4&0.727\\
		0.1  &0.8&0.726\\
	\bottomrule
	\end{tabular}
\end{table}

Finally, in Table \ref{table:new} we show the \eval{accuracy} for the SVM-based classifier for the \data{Scene} dataset for all the tested combinations of the learning rate and the momentum, in which the number of hidden units is fixed to $120$. The \eval{accuracy} for the \ECC\ (without RBM generated features) is $0.695$ and in this case any combination of learning rate and momentum does better, which indicates that with a sufficient number of hidden units, the RBM learning is quite robust and not overly sensitive to hyperparameter settings. 

\subsubsection{RAndom K labEL subsets}

RAkEL is a truncated power set method in which we try all combinations for 3 labels and we report an ensemble with $2L$ classifiers. We use the same hyperparameter setting as we did for the ECC to make the results comparable across multi-label classification paradigms, as reported in Table \ref{tab:c} and we report the \eval{acuracy} in Table \ref{table:1.accuracy}.  \\

\begin{table}[h]
	\centering
\caption{\label{table:1.accuracy} We report the \eval{accuracy} for RAkEL with and without feature extraction using RBMs using an SVM and a logistic regression based multi-label classifiers.}
		\begin{tabular}{rrrrr}
		\toprule
		 & \multicolumn{2}{c}{SVM} & \multicolumn{2}{c}{Log-Reg} \\
		 \cmidrule{2-3} \cmidrule{4-5}
								&\RAkr      &\RAk     	&\RAkr      &\RAk     \\
		\midrule
			  \data{Music}      &  0.581     & 0.579     &\textbf{0.538}      &0.465    \\
			  \data{Scene}      & \textbf{0.712}     & 0.684      &\textbf{0.663}      &0.469     \\
			  \data{Yeast}      & 0.537     & 0.537        &\textbf{0.497}      & DNF \\
			  \data{Genbase}    & 0.984     & 0.984      &0.968      &\textbf{0.976}    \\
			  \data{Medical}    &0.652     & \textbf{0.743}     &0.494      &\textbf{0.639}      \\
			  \data{Enron}      & \textbf{0.452}     & 0.413   &\textbf{0.376}      &0.273      \\
			  \data{Reuters}    & 0.342     & 0.337       &\textbf{0.285}      & DNF    \\
		\bottomrule
		\end{tabular}
	
\end{table}

The results for this paradigm are similar to the ones that we reported for the ECC in the previous section. For the logistic regression (a linear classifier) the RBM generated features lend themselves for accurate predictions when compared with the unprocessed features with the same baseline classifier and they are comparable to the results achieved for the nonlinear SVM classifier. After processing the features with an RBM we might not need to rely on a nonlinear classifier. For the SVM using the RBM generated features does not help, but it does not hurt either, in terms of \eval{accuracy}, as the SVM nonlinear mapping is versatile to learn any nonlinear mapping. 

\subsubsection{Pairwise Classification}

We implemented a pairwise approach, namely Four-class pairWise classifier (\texttt{FW}), in which we build models to learn classes $y_{jk} \in \{00,01,10,11\}$ for each label pair $1 \le j < k \le L$, dividing each into votes for the individual labels $\hat{y}_j$ and $\hat{y}_k$ and using a threshold at classification time. We find that overall it obtains better predictive performance than the pairwise methods that create decision boundaries between labels (where $y_{jk} \in \{01,10\}$), as in \cite{CLR}, for example, especially with SVMs. We report the \eval{accuracy} in Table \ref{table:2.accuracy}, using the same hyper parameters as we did for the ECC to make the results comparable across multi-label classification paradigms, as reported in Table \ref{tab:c}.\\

\begin{table}[h]
	\centering
\caption{\label{table:2.accuracy} We report the \eval{accuracy} for \texttt{FW} with and without feature extraction using RBMs, using an SVM and a logistic regression based multi-label classifiers.}
		\begin{tabular}{rrrrr}
		\toprule
		 & \multicolumn{2}{c}{SVM} & \multicolumn{2}{c}{Log-Reg} \\
		 \cmidrule{2-3} \cmidrule{4-5}
								&\FWr       &\FW       	&\FWr      &\FW     \\
		\midrule
			  \data{Music}      &   0.578     & 0.573     &\textbf{0.549}      &0.492    \\
			  \data{Scene}      & \textbf{0.694}     & 0.649      &\textbf{0.660}      &0.490     \\
			  \data{Yeast}      & 0.537     & 0.538       &0.507      &0.495 \\
			  \data{Genbase}    & 0.985     & 0.985      &0.949      &\textbf{0.975}   \\
			  \data{Medical}    &0.571     & \textbf{0.748}    &\textbf{0.492}      &DNF     \\
			  \data{Enron}      & \textbf{0.463}     & 0.408  &\textbf{0.376}      &DNF     \\
		\bottomrule
		\end{tabular}
\end{table}

The conclusions are similar to the other two paradigms. The linear classifier (logistic regression) does significantly better with the RBM generated features than with the original input space, while the SVM nonlinear classifier is versatile enough to provide accurate predictions with or without RBM generated features. Fortunately, the linear classifier with RBM generated features is quite close to the SVM-based classifier and allows to interpret which RBM features contribute to each label, hence we can provide intuitive interpretations for each RBM features, while it is hard to get such interpretation from the SVM nonlinear mapping.

\subsection{DBN performance}

\begin{figure}[h]
    \centering 
	\includegraphics[scale=1.0]{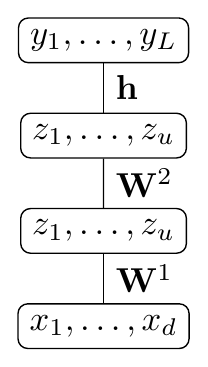}
	\caption{\label{fig:DBMh} A deep learning setup for multi-label classification.} 
\end{figure}

After analyzing the performance of the RBM generated features, we focus on two DBN structures for multi-label classification:
\begin{itemize}
	\item \DBNecc: a network of two hidden layers, the final of which is united with the labels in a new dataset and trained with \texttt{ECC} (see \fig{fig:DBNa})
	\item \DBNbpm: a network of three hidden layers where the final layer represents the labels; fine-tuned with back propagation (see \fig{fig:DBNb})
\end{itemize}
Both setups can be visualised in \fig{fig:DBMh}, where $\h\equiv\W^\ell$ in the case of \DBNbpm.

We use $u = d/5$ hidden units, 1000 RBM epochs, 100 BP epochs (on \DBNbpm), and the best of either $\alpha=0.8,\lambda=0.1$ and  $\alpha=0.8,\lambda=0.1$ on a 67:33 percent internal train/test validation (taking advantage of the fact, as we explained earlier, that the choice of learning rate and momentum is fairly robust given enough hidden units).

In Table \ref{table:1}, we compare the \eval{accuracy} for the proposed DBMs structures and the previously proposed methods. We have also added \MLkNN, \BPMLL, and \IBLR\ (see Section \ref{sec:prior} for details). In this table we can see that the \DBNecc\ is either the best classifier or close to the best, which give sense that the features generated by the second layer improve the first layer. For example, the only database (\data{Medical}) in which the \ECCr\ was not good enough compared to the \ECC\, now the \DBNecc\ and \DBNbpm\ do almost as good as \ECC\ and the performance on the other databases is also improved (or not degraded). This structure seems to be amenable for multi-label classification and competitive with all the proposed paradigms in the literature.\\

\begin{table*}
	\centering
{
    \caption{\label{table:1}Comparing multi-label methods under \eval{accuracy}. Highest results are set in \textbf{boldface}.}
	\begin{tabular}{rrrrrrrrrr}
		\toprule
                    &\DBNbpm       &\DBNecc   &\MLkNN    &\IBLR   &\ECCr    &\ECC      &\RAk    &\FW        &\BPMLL    \\
        \midrule                                                      
      \data{Music}  & 0.577        & \textbf{0.581}    &  0.542  & 0.545   & \textbf{0.581}   & 0.576    & 0.579  & 0.573      & 0.533    \\
      \data{Scene}  & 0.731        & \textbf{0.742}    & 0.696   & 0.697   & 0.731   & 0.710    & 0.684  & 0.649      & 0.552    \\
      \data{Yeast}  & 0.529        & 0.531    & 0.537   & \textbf{0.539}   & 0.532   & 0.535    & 0.537  & 0.538      & 0.491    \\
      \data{Genbase}& 0.984        & \textbf{0.985}    & 0.950   & 0.918   & 0.979   & 0.981    & 0.984  & \textbf{0.985}      & 0.049    \\
      \data{Medical}& 0.746        & 0.742    & 0.596   & 0.494   & 0.695   & \textbf{0.770}    & 0.743  & 0.748      & 0.053    \\
      \data{Enron}  & 0.442        & \textbf{0.480}    & 0.353   & 0.363   & 0.469   & 0.454    & 0.413  & 0.408      & 0.144    \\
      \data{Reuters}& 0.410        & 0.451    & 0.408   & 0.357   & 0.459   & \textbf{0.461}    & 0.337  & DNF        & 0.004    \\
		\bottomrule
	\end{tabular}
}
\end{table*}

\section{Conclusions}


Our empirical evaluation over a variety of multi-label datasets shows that a selection of high-performing multi-label methods from the literature can be improved upon by using an RBM-processed feature space. The labels become easier to model at training time, and predict at inference time. We obtained an improvement of up to $15$ percentage points in accuracy than when using the original feature space directly. Our study showed that important improvements can be obtained in multi-label classification with respect to both scalability and predictive performance when using deep learning in the area of multi-label classification. As a result, we can recommend to multi-labellers to focus more on feature modelling, rather than solely on modelling dependencies between the output labels. Our multi-label DBN models achieved the best predictive performance overall compared with seven competing methods from the multi-label literature.

\bibliographystyle{plain}
\bibliography{RBMs}

\end{document}